\documentclass{article}

\usepackage[final]{nips_2016}

\usepackage[utf8]{inputenc} % allow utf-8 input
\usepackage[T1]{fontenc}    % use 8-bit T1 fonts
\usepackage{hyperref}       % hyperlinks
\usepackage{url}            % simple URL typesetting
\usepackage{booktabs}       % professional-quality tables
\usepackage{amsfonts}       % blackboard math symbols
\usepackage{nicefrac}       % compact symbols for 1/2, etc.
\usepackage{microtype}      % microtypography
\usepackage[pdftex]{graphicx}
\usepackage[normalem]{ulem}

\title{Genetic Architect: Discovering Genomic Structure with Learned Neural Architectures}

\author{
  Laura Deming\thanks{Equal Contribution}, Sasha Targ\footnotemark[1], Nate Sauder, Diogo Almeida, Chun Jimmie Ye \\
  Institute for Human Genetics\\
  University of California,
  San Francisco, CA 94143, USA\\
  \\
  Enlitic\\
  San Francisco, CA 94111, USA\\  
 \texttt{ldeming.www@gmail.com, sasha.targ@ucsf.edu} \\
 \texttt{nate@enlitic.com, diogo@enlitic.com, jimmie.ye@ucsf.edu} \\
}

\begin{document}

\maketitle

\begin{abstract}
%\textbf{Abstract}
Each human genome is a 3 billion base pair set of encoding instructions. Decoding the genome using deep learning fundamentally differs from most tasks, as we do not know the full structure of the data and therefore cannot design architectures to suit it. As such, architectures that fit the structure of genomics should be learned not prescribed. Here, we develop a novel search algorithm, applicable across domains, that discovers an optimal architecture which simultaneously learns general genomic patterns and identifies the most important sequence motifs in predicting functional genomic outcomes. The architectures we find using this algorithm succeed at using only RNA expression data to predict gene regulatory structure, learn human-interpretable visualizations of key sequence motifs, and surpass state-of-the-art results on benchmark genomics challenges.
\end{abstract}

\section{Introduction}
\label{intro}
Deep learning demonstrates excellent performance on tasks in computer vision, text and many other fields. Most deep learning architectures consist of matrix operations composed with non-linearity activations. Critically, the problem domain governs how matrix weights are shared. In convolutional neural networks – dominant in image processing – translational equivariance (“edge/color detectors are useful everywhere”) is encoded through the use of the convolution operation; in recurrent networks – dominant in sequential data – temporal transitions are captured by shared hidden-to-hidden matrices. These architectures mirror human intuitions and priors on the structure of the underlying data. Genomics is an excellent domain to study how we might learn optimal architectures on poorly-understood data because while we have intuition that local patterns and long-range sequential dependencies affect genetic function, much structure remains to be discovered.

The genome is a very challenging data type, because although we have tens of thousands of whole genome sequences, we understand only a small subset of base pairs within each sequence. While the genetic code allows us to annotate the 5\% of the genome encoding proteins ($\sim$20,000 genes in the human genome), we do not have a “grammar” for decoding the rest of the non-coding sequences (90-95\% of the mouse and human genomes) important for gene regulation, evolution of species and susceptibility to diseases. The availability of a wealth of genomic assays (PBM, CHIP-seq, Hi-C) allows us to directly measure the function of specific regions of the genome, creating an enormous opportunity to decode non-coding sequences. However, the overwhelming volume of new data makes our job as decoders of the genome quite complex. The design and application of new domain-specific architectures to these datasets is a promising approach for automating interpretation of genomic information into forms that humans can grasp.

\section{Related Work}
Inspired by human foveal attention where global glances drive sequential local focus, attention components have been added to neural networks yielding state-of-the-art results on tasks as diverse as caption generation, machine translation, protein sublocalization, and differentiable programming. There are two main architectural implementations: hard attention, where the network's focus mechanism non-differentiably samples from the available input, and soft attention, where the component outputs an expected glimpse using a weighted average. Beyond biological inspiration, these components enable improved performance and excellent intepretability. Other techniques have been applied for interpreting neural networks without changing their architectures (\citet{simonyan2013deep}, \citet{zeiler2014visualizing}, \citet{springenberg2014striving}), but these are simply heuristics for finding the relevant regions of an input and do not work with all existing modern neural network components. 

Previous groups have demonstrated excellent progress applying deep learning to genomics. Both \citet{alipanahi2015predicting} and \citet{lanchantin2016deep} provide initial results on the task of learning which sequences a transcription factor (a biological entity which affects gene expression) can bind using convolutional architectures. This problem appears suited for convolution, as motifs determining binding are expected to be modular ($\sim$7-10 base pair units) and the setup of the task (preselected input sequences of fixed short length) does not allow for learning significant long-term dependencies. In particular, \citet{alipanahi2015predicting} demonstrated that a single-layer convolutional neural network, DeepBind, outperformed 26 other tested machine learning approaches in predicting probe intensities on protein binding microarrays from the DREAM5 PBM challenge, and then showed that the same architecture generalized to the related task of predicting transcription factor binding sites (TFBSs) from sequencing measurements of bound DNA. Subsequently, \citet{lanchantin2016deep} showed that a deeper network with the addition of highway layers improved on DeepBind results in the majority of cases tested  \citep{highway}. In addition, Basset \citep{kelley2015basset}, an architecture trained to predict motifs of accessible DNA from sequencing regions of open chromatin, was able to map half of the first layer convolutional filters to human TFBSs.

\section{Development of Genetic Architect}
Deep learning algorithm development is often dependent on the knowledge of human domain experts. Researchers in domains such as computer vision and natural language processing have spent much more time tuning architectures than in genomics. The challenge in genomics is that our insufficient understanding of biology limits our ability to inform architectural decisions based on data. Early genomic deep learning architectures have shown promising results but have undertaken only limited exploration of the architectural search space over possible components. In addition, not all components work well together, and there is evidence optimal component choice is highly dependent on the domain. Accordingly, we design a novel road-map for applying deep learning to data on which we have limited prior understanding, by developing an iterative architecture search over standard and cutting-edge neural net building blocks.

Prior approaches to architecture search focus on finding the best architecture in a single step, rather than sequentially learning more about the architecture space and iteratively improving models (\citet{bergstra2011algorithms}, \citet{bergstra2012random}, \citet{snoek2012practical}). Our framework understands the results allowing us to sequentially narrow the search space and learn about which combinations of components are most important. Since our algorithm limits the most important hyperparameters to their best ranges, they no longer dominate the search space and we discover additional hyperparameters that are most important and can help us create a highly tuned architecture. The sequential nature allows us to fork our architectural search into independent subspaces of coadapted components, thus enabling further search in each parallel branch to be exponentially more efficient than considering the union of all promising architectures.

\begin{figure}[!ht]
  \centering
  \includegraphics[scale=0.3]{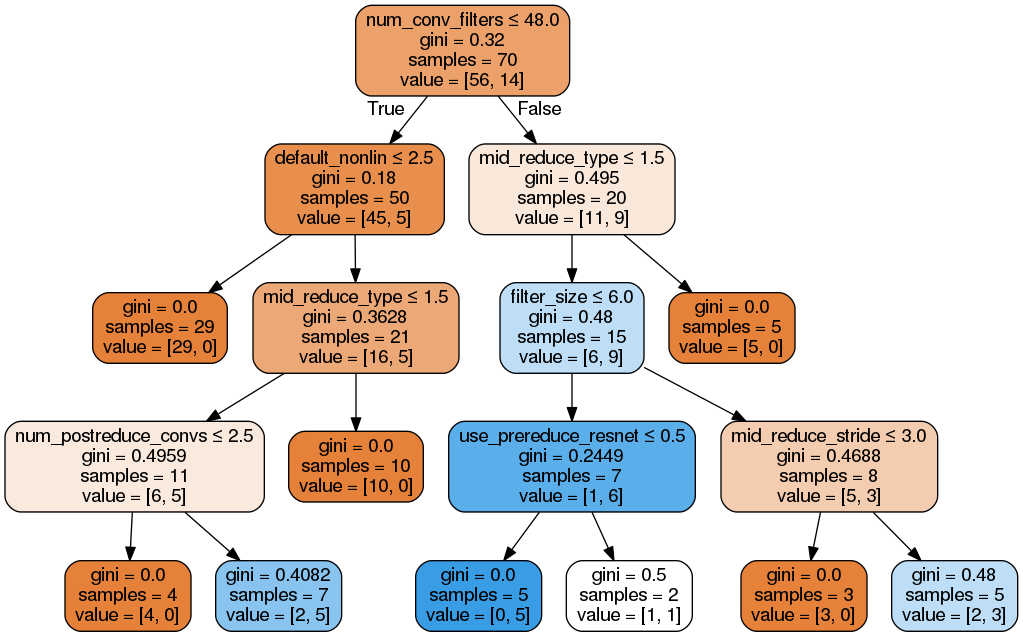}
    \caption{Example decision tree output from Genetic Architect visualization tool depicting significant hyperparameters for models with top 20\% performance.}
  \label{fig:favorite_tree}
\end{figure}

The heart of the framework is an interactive visualization tool (Figure \ref{fig:favorite_tree}). Given any hyperparameter optimization run, it produces common patterns for the best few datapoints and presents this information in highly-interpretable decision trees showing effective architectural subspace and plots of the interactions between the most significiant hyperparameters, informing general domain intuition and guiding future experiments. The framework is general enough to be applied to other domains, and is orthogonal to existing hyperparameter optimization algorithms. These algorithms can be applied in the inner loop of the sequential search of our tool, which then interprets the results and informs the user about the domain and how to manually prune the search space.

\begin{figure}[!ht]
  \centering
  \includegraphics[scale=0.4]{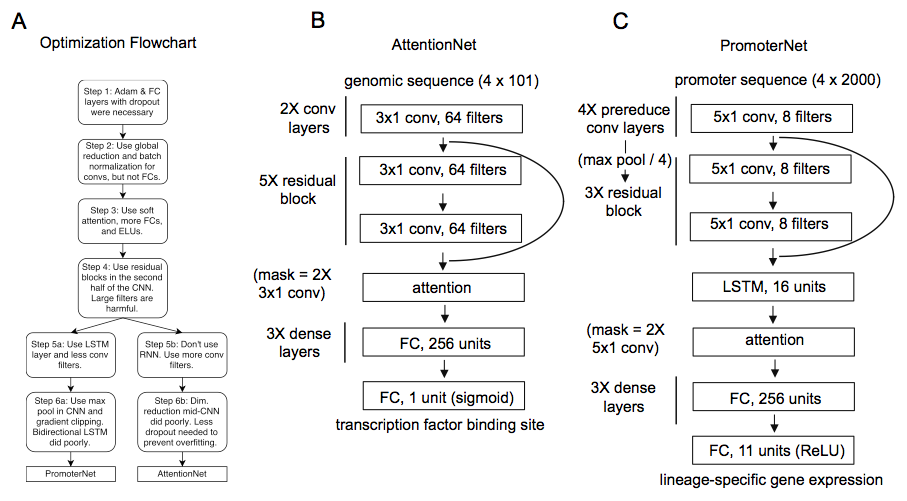}
    \caption{Schematic of hyperparameter optimization and final architecture designs. A) Overview
of steps taken in hyperparameter optimization to generate AttentionNet and PromoterNet. B)
AttentionNet architecture. C) PromoterNet architecture.}
  \label{fig:model_diagram}
\end{figure}

We employ Genetic Architect to discover an optimal architecture for a novel genome annotation task, regression to predict lineage-specific gene expression based on genomic sequence inputs, for which six stages of architecture search were required. Figure \ref{fig:model_diagram}A shows the sequential process of architecture search, the most important findings at each stage of the process, and tool-guided division of the search into two separate promising architectures. By splitting effective architectures into separate branches for further optimization, our tool identifies high-performing but architecture-specific choices that may be difficult to notice when architectures are mixed together.

The application of our tool demonstrates the power in refining architectural components that dominate results to uncover additional hyperparameter combinations that perform well together. Several examples we encounter during use of the tool for design of architectures for genomics follow: 1) removal of batch normalization demonstrated clear superiority of exponential linear units, 2) dimensionality reduction in the middle of the convolutional network module was beneficial to the recurrent-based architectures (perhaps since it decreased the distance of long-range dependencies), and 3) in contrast, non-recurrent architectures required wider layers (likely to enable processing of long-range dependencies in final dense layers). In our search over architectures using soft attention, we found that fully-connected layers were preferred to convolutional layers as it made processing global information more important. Finally, only by proceeding through several steps of optimization did we find the unintuitive result that bidirectional LSTMs did not help with attentional models (perhaps because the preceding layer effectively attends to a single location, making it difficult to combine information from both directions).

The final models learned by Genetic Architect consist of several initial layers of convolutions, residual blocks, an LSTM layer in the case of the PromoterNet architecture, and an attention-based dimensionality reducing step followed by fully-connected layers. Previous approaches to genome annotation use convolutional networks, which are ideal for detecting local features. However, more closely approximating the structure of genomic information would take into account that a real genome is a sequence, not a disjointed set, of local features – an input type on which recurrent architectures generally excel. In addition, with larger sequences to analyze (identifiable promoter sequences reach hundreds of base pairs in length), a neural network must learn to focus on the most important parts of the sequence and integrate new information derived from each part with the contextual information of the previously-seen sequence. As such, long genomic sequences seem an ideal fit for the recurrent attentional models learned.

\begin{figure}[!ht]
  \centering
  \includegraphics[scale=0.7]{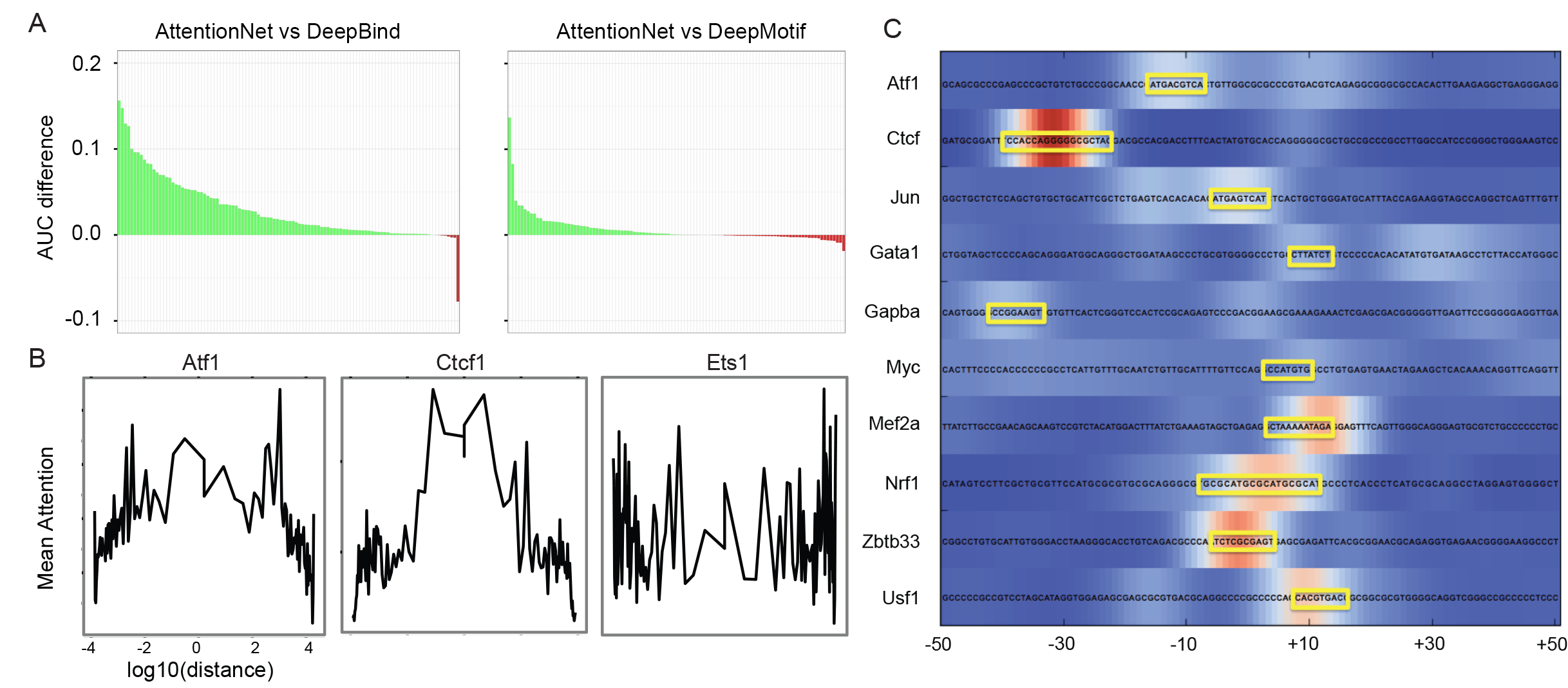}
    \caption{Results of AttentionNet on transcription factor binding site (TFBS) task. A) AttentionNet
models outperform DeepMotif and DeepBind models trained on corresponding datasets. Each bar represents the difference in AUC for one of 108 different datasets. B) Mean of attention mask over all sequences in experiment. C) Recovery of transcription factor motifs by visualization of attention masks produced by AttentionNet over example sequences.
}
  \label{fig:tfbs}
\end{figure}

\begin{figure}[!ht]
  \centering
  \includegraphics[scale=1.3]{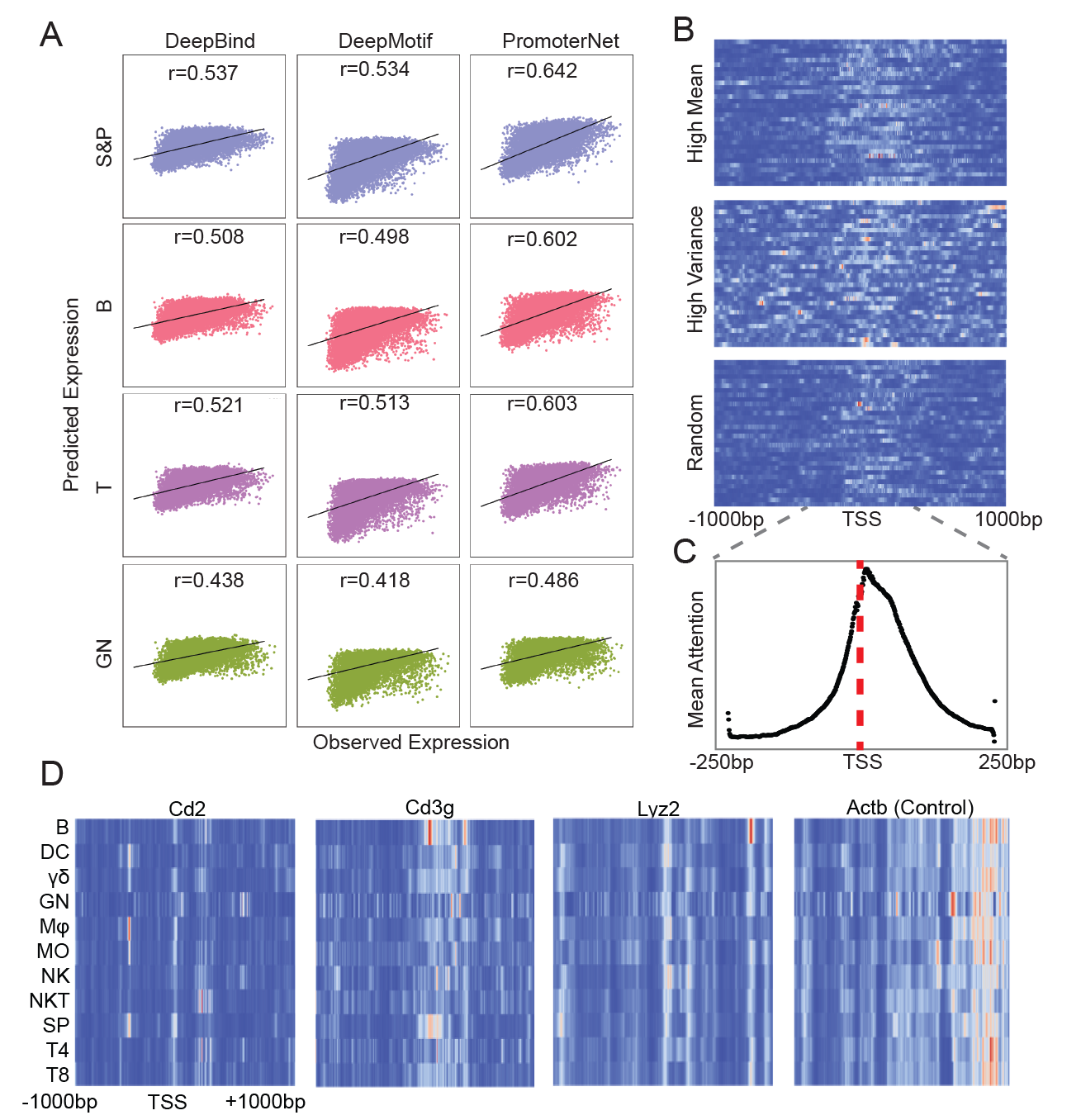}
  \caption{Results of PromoterNet on ImmGen lineage-specific expression prediction (ILSEP) task. A) Comparison of predicted versus observed gene expression for DeepBind, DeepMotif, and PromoterNet architectures. B) Visualization of attention mask over selected promoter sequences. C) Mean attention mask over all promoters. D) Visualization of attention masks learned by models trained on data from single lineages.}
  
  \label{fig:immgen}
\end{figure}

\begin{table}
  \caption{Mean and median AUC of models and percentage of datasets on which each model outperforms DeepBind or DeepMotif.}
  \label{table:tfbs-auc}
  \begin{center}
    \begin{tabular}{lrrrr}
      \toprule
      model & mean AUC & median AUC & vs DeepBind & vs DeepMotif \\
      \hline
      DeepBind & 0.904 & 0.932 & - & -\\
      DeepMotif & 0.927 & 0.949 & 85.2 & -\\
      AttentionNet & 0.933 & 0.952 & 92.6 & 67.6\\
      \bottomrule
    \end{tabular}
  \end{center}
\end{table}

\section{Experimental Results}

\subsection{Tasks}

\subsubsection{Transcription factor binding site (TFBS) classification}

The TFBS binary classification task was proposed by \citet{alipanahi2015predicting} and used as a benchmark by \citep{lanchantin2016deep}. The basic motivation is to learn a classifier that correctly predicts, from empirical binding data on a training sample of short DNA sequences, which sequences in a separate test set are TFBS (likely to be bound by biological entities, in this case, a given transcription factor protein).

The input and target data for the TFBS classification task consists of 108 datasets with an average of $\sim$31,000 sequences of 101 characters per dataset. Each sequence is a string of base pairs (A, C, G, or T) and is transformed into an array with one-hot encoding. Each sequence has an associated label (1 or 0) which indicates if this sequence is a TFBS. Each dataset represents a different chromatin immunoprecipitation sequencing (ChIP-seq) experiment with a specified transcription factor, and each sequence in the dataset a potential binding site. For each positive example, a negative example is generated. The data included in the TFBS classification task derive from ENCODE CHIP-seq experiments performed in K562 transformed human cell lines \citep{encode}.

\subsubsection{ImmGen lineage-specific expression prediction (ILSEP) regression}
In addition to the TFBS classification problem, neural network architectures could be extended to treat a much broader and complex variety of problems to do with interpreting biological data. Here, we develop a novel genomic benchmark task, ILSEP, which requires regression to predict empirically-determined related target data, namely, prediction of the amount of various biological entities produced in different cellular contexts given an input genomic sequence.

The input dataset for the ILSEP task is 14,116 one-hot encoded (4,2000) input promoter sequences and corresponding (243,) floating point gene expression outputs ranging between 2.60 and 13.95 (see appendix for details). We split the dataset using 10-fold cross validation to obtain predictions for all promoter gene expression pairs.

\subsection{Results on TFBS}

\subsubsection{Model performance}

We benchmark the performance of AttentionNet models learned by hyperparameter optimization described above against published state-of-the-art neural network models on the TFBS task, DeepBind \citep{alipanahi2015predicting} and DeepMotif \citep{lanchantin2016deep}. To compare the architectures, we train models for each of 108 datasets, as in \citet{lanchantin2016deep}. In a head-to-head comparison on each dataset, AttentionNet outperforms DeepMotif in 67.6\% of cases and the mean AUC across datasets for AttentionNet is 0.933, improving over both DeepMotif (0.927) and DeepBind (0.904) (Table \ref{table:tfbs-auc}).

\subsubsection{Prediction and visualization of genomic information}
Interpretable information about sequence features is an important consideration for genomic learning tasks where fundamental understanding of biology is as important as prediction power. We hypothesize that a net which performed well on the TFBS classification task would be able to make biologically meaningful inferences about the sequence structure. We show that the mean attention weights across all positive sequences show a distinct “footprint” of transcription factor (TF) binding consistent with known nucleotide preferences within each sequence (Figure \ref{fig:tfbs}B). Further, visualizing the attention mask (with the addition of Gaussian blur) across input sequences for 10 representative TFs showed the net focusing its attention on parts of the sequence known to be regulatory (Figure \ref{fig:tfbs}C).

To see if we could directly obtain motif sequences from the net, we took 10 nucleotides surrounding the position with highest attention for each of the top 100 sequences of a TF and averaged across the motifs. We took the maximum score for each nucleotide per position and queried the results against JASPAR, the ``gold standard'' TFBS database (with q < 0.5) \citep{jaspar}. 30/57 motifs possible to check (i.e. in JASPAR) were correct, and 39/57 corresponded to at least one transcription factor. By additionally searching the top 3 recurring sequences attended to for each TF, we recover a total of 42/57 correct motifs.

\subsection{Results on ILSEP}

\subsubsection{Model performance}

The PromoterNet architecture demonstrates a marked gain in performance over DeepBind and DeepMotif architectures adapted to the ILSEP regression task, achieving an average Pearson r correlation value of 0.587 between out-of-sample predictions and target expression values across lineages, compared to 0.506 and 0.441 for DeepBind \citep{alipanahi2015predicting} and DeepMotif \citep{lanchantin2016deep} respectively (Figure \ref{fig:immgen}A). We also train PromoterNet architectures on single task regression with a separate model for each of the 11 lineages and on cell type specific multi-task regression with one output unit for each of 243 cell types, which obtains similar improvements in average Pearson r correlation value of 0.592 over 0.502 for DeepBind and 0.498 for DeepMotif.

\subsubsection{Promoter element recovery and visualization of proximal regulatory elements}
Visualization of attention mask weights from the PromoterNet model reveals attended locations over promoter sequences of 32 genes selected for highest mean expression across lineages are enriched directly adjacent to the TSS, suggesting that properties of the core promoter sequence constitute the most informative features for genes that do not show differences in expression across lineages (Figure \ref{fig:immgen}B) (see appendix for list of genes). In contrast, attended locations over promoter sequences of 32 genes with maximal variance in expression across lineages span a much greater range of positions. This indicates that in genes with the greatest degree of lineage-specific expression, informative sequence features can occur throughout the promoter sequence. This observation merits follow up given previous reports that the performance of (non-deep) classifiers for cell type specific expression tasks trained only on TSS proximal promoter information is close to that of a random classifier \citep{promoter}. Consistent with accepted understanding that TSS proximal regions contain genomic elements that control gene expression levels, we observe the maximum of average attention mask weights across all promoters occurs at the center of input sequences, which corresponds to core promoter elements required for recruitment of transcriptional machinery \citep{maston} (Figure \ref{fig:immgen}C).

PromoterNet models trained for multi-task regression result in a global attention mask output across all lineages. To investigate whether the PromoterNet architecture is capable of learning distinct features for each lineage, we also visualize attention weights for a given promoter sequence from separate models, each trained on expression data for a single lineage. We find that genes selected for maximal variance in expression demonstrate distinct patterns of learned attention across lineages, while a shared pattern of attention is learned for a control gene with high mean expression in all lineages even when each lineage was trained on a separate model (Figure \ref{fig:immgen}D).

\section{Conclusion}
We tackle the problem of discovering architectures on datasets where human priors are not available. To do so we create a novel architecture search framework that is domain agnostic, is capable of sequential architectural subspace refinement and informing domain understanding, and is composable with existing hyperparameter optimization schemes. Using this search algorithm, we create state-of-the art architectures on significant challenges in the domain of genomics utilizing a combination of standard and cutting-edge components. In particular, the learned architecture is capable of simultaneous discovery of local and non-local patterns, important subsequences, and sequential composition thereby capturing substantial genomic structure.

\section{Appendix}
\begin{table}
  \caption{Search space explored for AttentionNet and PromoterNet architectures, including techniques
    from \citet{maas2013rectifier}, \citet{graham2014spatially}, \citet{shah2016deep}, \citet{ioffe2015batch}, \citet{resnet}, \citet{hochreiter1997long}, \citet{kingma2014adam}, \citet{sutskever2013importance}, \citet{dropout}.}
  \label{table:hyperparams}
  \begin{center}
    \begin{tabular}{ll}
      Hyperparameter & Values\\
      \hline
      conv filter size & 3, 5, 7, 9\\
      nonlinearity & ReLU, Leaky ReLU, Very Leaky ReLU, ELU\\
      using batch normalization for convs & True, False\\
      number of conv filters & 8, 16, 32, 64\\
      number of convs before dim. reduction & 1, 2, 3, 4, 5\\
      using residual blocks before dim. reduction & True, False\\
      type of dim. reduction & None, Max Pool, Mean Pool, Strided Conv\\
      dim. reduction stride & 2, 4, 8\\
      number of convs after dim. reduction & 1, 2, 3, 4, 5\\
      using residual blocks after dim. reduction & True, False\\
      number of RNN layers & 0, 1, 2\\
      number of units in RNN & 16, 32, 64\\
      RNN type & Simple RNN, LSTM\\
      using bidirectional RNNs & True, False\\
      RNN gradient clipping & 0, 1, 5, 15\\
      global reduction type & None, Attention, Max Pool, Mean Pool\\
      number of FC layers & 0, 1, 2, 3\\
      number of units in FC & 64, 128, 256, 512, 1024\\
      using batch normalization for FCs & True, False\\
      dropout probability for FCs (after) & 0., 0.1, 0.2, 0.3, 0.4, 0.5, 0.6\\
      L2 regularization & 0, 1e-3, 1e-4, 1e-5\\
      optimizer & Adam, SGD w/ Nesterov Momentum\\
      learning rate scale & 0.03, 0.1, 0.3, 1.0, 2.0, 10.\\
      batch size & 25, 50, 125, 250\\
    \end{tabular}
  \end{center}
\end{table}

\begin{table}
  \caption{AttentionNet architecture}
  \label{table:attentionnetb}
  \begin{center}
    \begin{tabular}{l}
      \toprule
      Layer\\
      \hline
      3x1 conv 64 filters + BN + ELU\\
      3x1 conv 64 filters + BN + ELU\\
      residual block (w/ 3x1 conv 64 filters + BN + ELU + 3x1 conv 64 filters + BN)\\
      residual block (w/ 3x1 conv 64 filters + BN + ELU + 3x1 conv 64 filters + BN)\\
      residual block (w/ 3x1 conv 64 filters + BN + ELU + 3x1 conv 64 filters + BN)\\
      residual block (w/ 3x1 conv 64 filters + BN + ELU + 3x1 conv 64 filters + BN)\\
      residual block (w/ 3x1 conv 64 filters + BN + ELU + 3x1 conv 64 filters + BN)\\
      attention (w/ 3x1 conv 64 filters + BN + tanh + 3x1 conv 1 filter + BN + softmax)\\
      FC 256 units + ELU + 0.2 dropout\\
      FC 256 units + ELU + 0.2 dropout\\
      FC 256 units + ELU + 0.2 dropout\\
      FC 1 unit + sigmoid\\
      \bottomrule
    \end{tabular}
  \end{center}
\end{table}

\begin{table}
  \caption{PromoterNet architecture}
  \label{table:promoternet}
  \begin{center}
    \begin{tabular}{l}
      \toprule
      Layer\\
      \hline
      5x1 conv 8 filters + BN + ELU\\
      5x1 conv 8 filters + BN + ELU\\
      5x1 conv 8 filters + BN + ELU\\
      5x1 conv 8 filters + BN + ELU\\
      4x1 maxpool, stride 4\\
      residual block (w/ 5x1 conv 8 filters + BN + ELU + 5x1 conv 8 filters + BN)\\
      residual block (w/ 5x1 conv 8 filters + BN + ELU + 5x1 conv 8 filters + BN)\\
      residual block (w/ 5x1 conv 8 filters + BN + ELU + 5x1 conv 8 filters + BN)\\
      LSTM 16 units\\
      attention (w/ 5x1 conv 8 filters + BN + tanh + 5x1 conv 1 filter + BN + softmax)\\
      FC 256 units + ELU + 0.3 dropout\\
      FC 256 units + ELU + 0.3 dropout\\
      FC 256 units + ELU + 0.3 dropout\\
      FC 1 unit + sigmoid\\
      \bottomrule
    \end{tabular}
  \end{center}
\end{table}

\subsection{ILSEP data processing}
For the input sequences in the ILSEP task, we use sequences spanning 1 kilobase upstream and downstream of transcriptional start sites (TSS), the region of the promoter at which production of the gene is initiated, for 17,565 genes from the Eukaryotic Promoter Database \citep{epd}. For the corresponding labels, we obtain expression data (log2 normalized microarray intensities) for each of these genes in each of 243 immune cell types from the ImmGen Consortium April 2012 release, which contains data for 21,755 genes x 243 immune cell types after quality control according to published methods \citep{immgen_qc}. Intersection of these two datasets by gene results in a dataset of 14,116 input promoter sequences and expression value target pairs.

To create lineage-specific gene expression value targets, we combine cell types into 11 groups following the lineage tree outlined in previous work: B cells (B), dendritic cells (DC), gamma delta T cells ($\gamma\delta$), granulocytes (GN), macrophages (M$\phi$), monocytes (MO), natural killer cells (NK), stem and progenitor cells (SP), CD4+ T cells (T4), and CD8+ T cells (T8), and average expression values across samples within each group \citep{ontogenet}.

\subsection{Selected genes (Figure \ref{fig:immgen}B)}
Highest mean expression across lineages: Rac2, Rpl28, Pfn1, Rpl9, Ucp2, Tmsb4x, Tpt1, Rplp1, Hspa8, Srgn, Rpl27a, Rpl13a, Cd53, Eef2, Rps26, Cfl1, Ppia, Gm9104, Rps2, Rps27, Actg1, Laptm5, Rpl21, Eef1a1, Rplp0, Gm15427, Pabpc1, B2m, Gapdh, Actb, Rpl17, Rps6

Highest variance in expression across lineages: Plbd1, Tlr13, Tyrobp, Ifitm2, Pld4, Pla2g7, Gda, Cd96, Gzma, Nkg7, Ctsh, Klrb1c, Ccl6, Prkcq, Itgam, Sfpi1, Itk, Ms4a4b, Alox5ap, Ly86, Cd2, Fcer1g, Gimap3, Il2rb, Gimap4, Ifitm6, Cybb, Ifitm3, Mpeg1, H2-Aa, Cd3g, Lyz2

Random control: Krt84, Lrrc8b, 8030411F24Rik, Syngr2, Spint3, Slc17a4, Slc22a23, Thoc6, AF529169, Phf5a, Yif1b, 4930467E23Rik, Pgam1, Pcdhb1, Bak1, Neu3, Plcb2, Fabp4, Srgap1, Olfr1339, Sox12, Atg7, Gdf10, 1810008A18Rik, 1700011A15Rik, Anks4b, Magea2, Pygb, Spc25, Rras2, Slc28a3, 9130023H24Rik

\bibliographystyle{unsrtnat}
\bibliography{REFERENCES}

\end{document}